# Variable importance in binary regression trees and forests

**Hemant Ishwaran**[*]

*Department of Quantitative Health Sciences*
*Cleveland Clinic, 9500 Euclid Avenue, Cleveland, OH 44195*
*e-mail:* `hemant.ishwaran@gmail.com`

**Abstract:** We characterize and study variable importance (VIMP) and pairwise variable associations in binary regression trees. A key component involves the node mean squared error for a quantity we refer to as a maximal subtree. The theory naturally extends from single trees to ensembles of trees and applies to methods like random forests. This is useful because while importance values from random forests are used to screen variables, for example they are used to filter high throughput genomic data in Bioinformatics, very little theory exists about their properties.

## Contents

## 1. Introduction

A binary recursively grown tree, $T$, can be thought of as a map that uniquely identifies a multivariable covariate $\mathbf{x} \in \mathscr{X}$ with tree node membership. Here $\mathbf{x}$ is a $d$-dimensional vector $\mathbf{x} = (x_1, \ldots, x_v) \in \mathbb{R}^d$, where $x_v$ are real valued variables. If $T$ has $M$ terminal nodes, then $T$ is the function $T : \mathscr{X} \to \{1, \ldots, M\}$ defined via

$$T(\mathbf{x}) = \sum_{m=1}^{M} mB_m(\mathbf{x}). \tag{1.1}$$

[*]Research partially supported by National Institutes of Health RO1 grant HL-072771.

Here $B_m(\mathbf{x})$ are 0-1 basis functions that, because of the recursive nature of $T$, partition $\mathscr{X}$.

An important class of binary trees are those associated with Classification and Regression Trees (CART), a widely used method in regression and multi-class data settings [1]. In CART, a binary tree is grown using recursive splitting based on node impurity. The best split for a node is found by searching over all possible binary splits for each variable, and identifying the split for a variable leading to the greatest reduction in tree node impurity. Binary trees are also implicitly used in Random Forests [2], a popular ensemble technique used for prediction. Rather than using a single tree, random forests constructs an ensemble predictor by averaging over a collection of binary trees. Each tree is grown from an independent bootstrap sample of the data, where at each node of the tree a randomly selected subset of variables of size *mtry* are chosen as candidate variables to split on. Averaging over trees, in combination with the randomization used in growing a tree, enables random forests to approximate a rich class of functions while maintaining low generalization error. This enables random forests to adapt to the data, automatically fitting higher order interactions and non-linear effects, while at the same time keeping overfitting in check. This has led to a great interest in the method and applications in many fields.

While CART and random forests are often used for exploratory data analysis, they can also be used to select variables and reduce dimensionality. This is done by ranking variables by some measure of importance and removing those variables with low rank. Variable importance (VIMP) was originally defined in CART using a measure involving surrogate variables (see Chapter 5 of [1]). Definitions in terms of mean overall improvement in node impurity for a tree have also been proposed. In regression trees, node impurity is measured by mean squared error, whereas in classification problems, the Gini index is used [3]. The most popular VIMP method to date, however, adopts a prediction error approach involving "noising-up" a variable. In random forests, for example, VIMP for a variable $x_v$ is the difference between prediction error when $x_v$ is noised up by permuting its value randomly, compared to prediction error under the original predictor [2, 4, 5, 6]. This type of definition has become quite popular. For example, there is a growing trend in Bioinformatics where importance values from forests are used as a filtering method to reduce dimensionality in high-throughput genomic data [7, 8, 9, 10].

Although VIMP has been studied in general settings [11], there is still a great need for a systematic theoretical study focusing on trees. In this paper we provide such a treatment for the class of binary regression trees and their forests. This applies to data settings where the outcome $Y$ is continuous. We define rigorously VIMP as well as a measure of association between pairs of variables, and derive theoretical properties for such values. We find that something we call a maximal subtree and its node mean squared error play a fundamental role in our results.

It should be noted carefully that the definition for VIMP adopted here differs from that used in random forests. The permutation method used by random forests is complex and proved too difficult to analyze theoretically in any detail. This motivated us to look for an approach simpler to study, but with similar key features. Hence, we have adopted a simpler definition whereby VIMP is defined as the difference between prediction error under noising and without noising, where noising up a variable corresponds to what can be thought of as a random left-right daughter node assignment. Section 3 provides a rigorous definition of this process and discusses in detail how this method compares with random forests.

## 2. Binary regression trees

We begin by first defining a binary regression tree.

DEFINITION 1. Let $\mathscr{L} = \{(Y_i, \mathbf{x}_i) : i = 1, \ldots, n\}$ denote the learning data. A binary tree $T$ grown from $\mathscr{L}$ is a tree grown using successive recursive binary splits on variables of the form $x_v \leq c$ and $x_v > c$ where split values $c$ are chosen from the observed values of $\mathbf{x}_i$. The terminal values $a_m$ for a terminal node of $T$ is the average of those $Y_i$ with $\mathbf{x}_i$ having node membership $m$. In other words, $a_m = \sum_{i=1}^n I\{T(\mathbf{x}_i) \in m\} Y_i / \sum_{i=1}^n I\{T(\mathbf{x}_i) \in m\}$.

A binary regression tree (hereafter simply refered to as a binary tree) must be of the form (1.1). Moreover, because of the nature of recursive partitioning, the basis functions $B_m(\mathbf{x})$ in $T$ are product splines of the form:

$$B_m(\mathbf{x}) = \prod_{l=1}^{L_m} \bigl[x_{l(m)} - c_{l,m}\bigr]_{s_{l,m}}.$$

Here $L_m$ are the number of splits used to define $B_m(\mathbf{x})$. Each split $l$ involves a variable, denoted by $x_{l(m)}$ for the $l(m)$-th coordinate of $\mathbf{x}$, and a splitting value for this variable denoted by $c_{l,m}$. The $s_{l,m}$ are $\pm 1$ binary values, and for any given $x$, $[x]_{+1} = 1\{x > 0\}$ and $[x]_{-1} = 1\{x \leq 0\}$. One can think of $B_m(\mathbf{x})$ as a 0-1 function that is 1 only if $\mathbf{x}$ sits inside the hyper-rectangular region defined by $x_{l(m)}$ and the splits $c_{l,m}$.

As mentioned in the Introduction, recursive binary partitioning ensures that the resulting decomposition of $\mathscr{X}$ induced by the basis functions is a non-overlapping partition. Hence, an important property possessed by $B_m(\mathbf{x})$ is orthogonality. Thus,

$$B_m(\mathbf{x}) B_{m'}(\mathbf{x}) = 0 \quad \text{if } m \neq m'. \tag{2.1}$$

This property will play a key role in our theoretical development.

## 3. A surrogate VIMP

We define the VIMP for a variable $x_v$ as the difference between prediction error when $x_v$ is "noised up" versus the prediction error otherwise. To noise up $x_v$ we adopt the following convention. To assign a terminal value to a case **x**, drop **x** down $T$ and follow its path until either a terminal node is reached or a node with a split depending upon $x_v$ is reached. In the latter case choose the right or left daughter of the node with equal probability. Now continue down the tree, randomly choosing right and left daughter nodes whenever a split is encountered (whether the split depends upon $x_v$ or not) until reaching a terminal node. Assign **x** the node membership of this terminal node.

The left-right random daughter assignment is designed to induce a random tree that promotes poor terminal values for cases that pass through nodes that split on $x_v$. In fact, one can simply think of the noising up process equivalently in terms of this random tree, which we shall denote by $T_v$. The prediction performance of $T_v$ is intimately tied to the VIMP of $x_v$, which is directly related to the location of $x_v$ splits in $T$. The more informative $x_v$ is, the higher up $x_v$ splits will appear in $T$, and consequently the more perturbed $T_v$ becomes relative to $T$. This results in a loss of prediction accuracy for $T_v$ and a high VIMP for $x_v$.

### 3.1. Comparison to random forests

While our noising up procedure is different than the current method employed in random forests, our procedure is designed to approximately mimic what is seen in practice. In random forests, VIMP is calculated by noising up a variable by randomly permuting it. A given $x_v$ is randomly permuted in the out-of-bag (OOB) data (the data not selected by bootstrapping) and the noised up OOB data is dropped down the tree grown from the in-bag data (bootstrap data). This is done for each tree in the forest and an out-of-bag estimate of prediction error is computed from the resulting predictor. The difference between this value and the out-of-bag error without random permutation is the VIMP of $x_v$. Large positive values indicate $x_v$ is predictive (since noising up $x_v$ increases prediction error), whereas zero or negative importance values identify variables not predictive [2].

Often one finds in random forests that variables that tend to split close to the root have a strong effect on prediction accuracy [12]. When such variables are randomly permuted, each tree is perturbed substantially, so when out-of-bag data is dropped down the tree, the resulting predictor has poor prediction. Averaging over such predictors results in an ensemble with high error rate and a large VIMP for the variable. Variables tending to split lower down in a tree, on the other hand, have much less impact, because as one travels through a binary tree the amount of data drops off exponentially fast. Consequently there are much fewer observations affected and the noised up predictor does not suffer so much.

Our left-right noising up process shares with random forests the key feature that VIMP is directly impacted by location of the primary split relative to the root node. For this reason we believe our approach can provide insight into VIMP for random forests.

However, the analogy is by no means perfect. A nice example of this, suggested by one of the reviewers of the paper, is as follows. Recall that at each node, while growing a tree, random forests randomly selects a subset of *mtry* variables to split on. In a problem with a large number of non-informative variables, it is possible that early on in the tree growing process the *mtry* variables for a node are all non-informative. Hence, the eventual split for the node is on a non-informative variable, high up in the tree. Trees are, however, highly data adaptive and can recover from scenarios like this. Splits further down the tree can be on informative variables and the tree can still be predictive. Now consider what happens when we drop a case down the tree using the left-right noising up process. If the case passes through the node split on the non-informative variable $x_v$, then it follows a random left-right path through potentially predictive splits. Since the node is high in the tree, the resulting random tree $T_v$ suffers in terms of prediction and $x_v$ can appear informative.

This type of scenario shows that a non-informative variable can appear informative over a single tree under our noising up process. However, over a forest such an effect will most likely be washed out. With many non-informative variables, the probability that $x_v$ splits high in a tree is small. Since the event has low probability, the forest will contain few trees with high $x_v$ splits. Averaging over trees pushes the VIMP of $x_v$ to zero. Moreover, for a single tree, this kind of problem can be resolved by slightly modifying the noising up process. Rather than using random left-right assignments on all nodes beneath $x_v$, use random assignments for only those nodes that split on $x_v$. This will impact prediction only when $x_v$ is informative and not affect prediction for non-informative variables. This modified process, in fact, was used in early attempts to study this problem. Unfortunately, though, while it was our feeling this type of VIMP could compete against the one used by random forests, it was still too complex for the kind of detailed theoretical development we were after.

It is important to emphasize that while deficiencies like those mentioned suggest our noising up process may not be practicable in all data examples, this does not detract from theory developed under this mechanism. As we define shortly, a key theoretical concept in our framework is the notion of a maximal subtree. We believe this is key to understanding VIMP, even complicated values like those used by random forests.

### *3.2. Subtrees*

To discuss maximal subtrees, first note the predictor associated with the original tree $T$ can easily be written in closed form. Denote the predictor by $\hat{\mu}(\mathbf{x})$. It

follows that

$$\hat{\mu}(\mathbf{x}) = \sum_{m=1}^{M} a_m B_m(\mathbf{x}).$$

Surprisingly, it is also possible to give an explicit representation for the predictor of $T_v$. To do so, we introduce the notion of a subtree and a maximal subtree.

DEFINITION 2. Call $\tilde{T}_v$ a $v$-subtree of $T$ if the root node of $\tilde{T}_v$ has daughters that depend upon an $x_v$ split. Call $\tilde{T}_v$ a maximal $v$-subtree if $\tilde{T}_v$ is not a subtree of a larger $v$-subtree (see Figure 1). For example, the root node tree $T$ is always a maximal $v$-subtree for some $v$.

Suppose that $T$ contains $K_v$ distinct maximal $v$-subtrees, $\tilde{T}_{1,v}, \ldots, \tilde{T}_{K_v,v}$. Let $M_{k,v}$ be the set of terminal nodes for the subtree $\tilde{T}_{k,v}$ and define $M_v = \bigcup_{k=1}^{K_v} M_{k,v}$. Observe that $M_{k,v}$ are disjoint sets and that $M_v$ is the set of terminal nodes of $T$ that can only be reached by going through at least one node involving a split on $x_v$. The following key Lemma shows that the predictor for $T_v$ can be written as a deterministic component (conditioned on $T$) depending on terminal nodes not in $M_v$ and a random component involving terminal nodes in $M_v$.

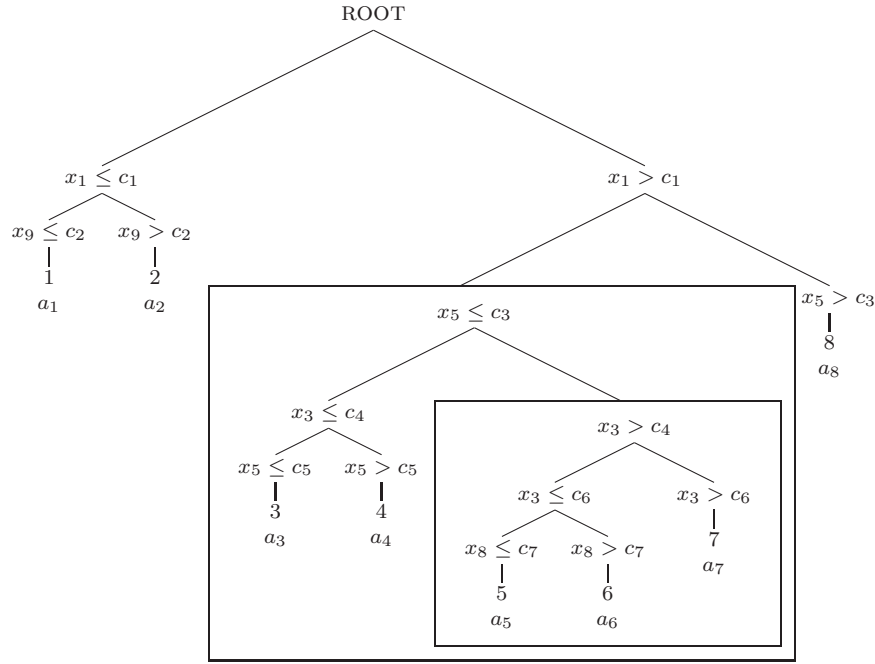

FIG 1. *Large box contains a maximal $v$-subtree, for $v = 3$. Smaller box containing tree with nodes $5, 6$ and $7$ is also a $v$-subtree, but it is not maximal.*

**Lemma 1.** *Let $\hat{\mu}_v(\mathbf{x})$ denote the predictor for $T_v$. Then,*

$$\hat{\mu}_v(\mathbf{x}) = \sum_{m \notin M_v} a_m B_m(\mathbf{x}) + \sum_{k=1}^{K_v} \tilde{a}_{k,v} I\{T(\mathbf{x}) \in M_{k,v}\}, \tag{3.1}$$

*where $\tilde{a}_{k,v}$ is the random terminal value assigned by $\tilde{T}_{k,v}$ under a random left-right path through $\tilde{T}_{k,v}$. Write $\tilde{P}_{k,v}$ for the distribution of $\tilde{a}_{k,v}$. Note that $\tilde{P}_{k,v}$ is conditional on the original tree $T$.*

PROOF. If $T(\mathbf{x}) \notin M_v$, then $\mathbf{x}$ is assigned the same terminal node as in $T(\mathbf{x})$. This explains the first sum in the expression (3.1). On the other hand if $T(\mathbf{x}) \in M_v$, then $\mathbf{x}$ passes through a root node of a maximal subtree, say $\tilde{T}_{k,v}$. Upon entering this node, $\mathbf{x}$ follows a random path through $\tilde{T}_{k,v}$ until eventually terminating at a node $m \in M_{k,v}$. Thus, the terminal value for $\mathbf{x}$ is the random value $\tilde{a}_{k,v}$ assigned by a random left-right path through $\tilde{T}_{k,v}$. □

### 3.3. Prediction error

To formally discuss VIMP it is necessary to first define what we mean by prediction error (PE). Let $g$ be some loss function. The PE for $\hat{\mu}$ is defined as

$$\mathscr{P}(\hat{\mu}) = \mathbb{E}\big(g(Y, \hat{\mu}(\mathbf{x}))\big).$$

The expectation on the right-hand side is joint over the original learning data $\mathscr{L}$ as well as the independent test data $(Y, \mathbf{x})$. It is assumed that $Y$ can be written as a regression model

$$Y = \mu(\mathbf{x}) + \varepsilon, \tag{3.2}$$

where $\varepsilon$ has zero mean and variance $\sigma^2 > 0$ and $\varepsilon$ is assumed independent of $\mu(\mathbf{x})$. In a similar way one can define PE for $\hat{\mu}_v$:

$$\mathscr{P}(\hat{\mu}_v) = \mathbb{E}\big(g(Y, \hat{\mu}_v(\mathbf{x}))\big).$$

Note the expection on the right-hand side involves an extra level of integration over $\tilde{a}_{k,v}$ via its distribution $\tilde{P}_{k,v}$.

We now define the VIMP of $x_v$. In this paper we confine attention to the case when $g(Y, \hat{\mu}) = (Y - \hat{\mu})^2$, corresponding to $L_2$-loss. The VIMP value for $x_v$ under $L_2$-loss is defined as

$$\Delta_v = \mathscr{P}(\hat{\mu}_v) - \mathscr{P}(\hat{\mu}).$$

Using standard arguments it is easy to show that under (3.2),

$$\mathscr{P}(\hat{\mu}) = \sigma^2 + \mathbb{E}(\mu(\mathbf{x}) - \hat{\mu}(\mathbf{x}))^2. \tag{3.3}$$

Now observe by Lemma 1 that $\hat{\mu}_v(\mathbf{x})$ can be rewritten as

$$\hat{\mu}_v(\mathbf{x}) = \hat{\mu}(\mathbf{x}) + \sum_{k=1}^{K_v} \sum_{m \in M_{k,v}} (\tilde{a}_{k,v} - a_m) B_m(\mathbf{x}).$$

Using a similar expansion for $\mathscr{P}(\hat{\mu}_v)$ as (3.3), deduce that the VIMP for $x_v$ is

$$\Delta_v = \mathbb{E}\left(R_v(\mathbf{x})^2\right) - 2\mathbb{E}\left\{R_v(\mathbf{x})\left[\mu(\mathbf{x}) - \hat{\mu}(\mathbf{x})\right]\right\}$$

where

$$R_v(\mathbf{x}) = \sum_{k=1}^{K_v} \sum_{m \in M_{k,v}} (\tilde{a}_{k,v} - a_m) B_m(\mathbf{x}).$$

Clearly $\Delta_v$ depends heavily upon the relationship between $a_m$ and $\tilde{a}_{k,v}$. For example, $\Delta_v = 0$ if $\tilde{a}_{k,v} = a_m$ for each $m \in M_{k,v}$ and each $k$. In this case, all terminal values within a maximal subtree are equal and randomly assigning a terminal value cannot perturb matters. Hence, $x_v$ is non-informative as further clustering of the data is ineffective. In general, however, noising up $x_v$ will perturb $\tilde{a}_{m,v}$ relative to $a_m$ and result in a positive or negative value for $\Delta_v$. Trying to detangle the resulting properties of $\Delta_v$ in such settings requires a theoretical framework which we discuss next.

## 4. A theoretical framework for VIMP

In order to study $\Delta_v$ systematically, it will be necessary to make a simplifying assumption about the true signal $\mu(\mathbf{x})$. In particular, it will be helpful if we assume that $\hat{\mu}(\mathbf{x})$ is a good approximation to $\mu(\mathbf{x})$. We shall assume the true signal has a similar topology to $T$:

$$\mu(\mathbf{x}) = \sum_{m=1}^{M} a_{m,0} B_m(\mathbf{x}), \tag{4.1}$$

where $a_{m,0}$ are the true, but unknown, terminal values. Observe that $\mu(\mathbf{x})$ is random because $B_m(\mathbf{x})$ are basis functions determined from the learning data. Nevertheless, we shall still think of $\mu(\mathbf{x})$ as our "population parameter" because the terminal values for $\mu(\mathbf{x})$ are fixed and do not depend upon the data. Note importantly that (3.3) still holds even though $\mu(\mathbf{x})$ depends upon the data.

**Remark 1.** Assumption (4.1) might seem simplistic at first glance, but we'll show later in Section 5 that it is quite realistic when put in the context of forests.

By exploiting the orthogonality property (2.1) of the tree basis functions, we can greatly simplify the expression for $\Delta_v$ under assumption (4.1). Consider the following theorem.

**Theorem 1.** *Assume the true model is (3.2) and that (4.1) holds (for any given $\mathscr{L}$). Then,*

$$\Delta_v = \mathbb{E}\left\{\sum_{k=1}^{K_v} \sum_{m \in M_{k,v}} \pi_m \left[s_{k,v}^2 + \left(\overline{a}_{k,v} - a_m\right)\left(\overline{a}_{k,v} + a_m - 2a_{m,0}\right)\right]\right\}, \quad (4.2)$$

*where $\pi_m = \mathbb{P}\{B_m(\mathbf{x}) = 1|\mathscr{L}\}$ and $s_{k,v}^2$ and $\overline{a}_{k,v}$ denote the variance and mean for $\tilde{a}_{k,v}$ under $\tilde{P}_{k,v}$.*

PROOF. Using orthogonality (2.1) and assumption (4.1), collect together common expressions to deduce that

$$\Delta_v = \mathbb{E}\left\{\sum_{k=1}^{K_v} \sum_{m \in M_{k,v}} \left[r_{k,m}^2 - 2r_{k,m}r_m\right]B_m(\mathbf{x})\right\},$$

where $r_{k,m} = \tilde{a}_{k,v} - a_m$ and $r_m = a_{m,0} - a_m$. The expectation sitting outside the sum can be partially simplified by integrating over $\tilde{P}_{k,v}$ while conditioning on $\mathscr{L}$ and the test data $\mathbf{x}$. A little bit of re-arrangement yields

$$\Delta_v = \mathbb{E}\left\{\sum_{k=1}^{K_v} \sum_{m \in M_{k,v}} \left[s_{k,v}^2 + \left(\overline{a}_{k,v} - a_m\right)\left(\overline{a}_{k,v} + a_m - 2a_{m,0}\right)\right]B_m(\mathbf{x})\right\}.$$

Continuing to keep $\mathscr{L}$ fixed, bring the expectation over $\mathbf{x}$ inside the sum. Only the term $B_m(\mathbf{x})$ depends upon $\mathbf{x}$, from which (4.2) follows. □

Theorem 1 has an interesting implication. Let $\tilde{P}_{k,v,0}$ be the distribution for $\tilde{a}_{k,v,0}$: the random terminal node assigned from a random path through the maximal subtree $\tilde{T}_{k,v}$ with terminal nodes replaced by $\{a_{m,0} : m \in M_{k,v}\}$. Call the quantity

$$\theta_0(k,v) = \sum_{m \in M_{k,v}} \pi_m \tilde{P}_{k,v,0}(\tilde{a}_{k,v,0} - a_{m,0})^2, \quad \text{for } k = 1, \ldots, K_v, \quad (4.3)$$

the *node mean squared error for the subtree* $\tilde{T}_{k,v}$. An immediate corollary to Theorem 1 is that the closer $\hat{\mu}(\mathbf{x})$ is to the true signal, the greater the effect node mean squared error has on VIMP.

**Corollary 1.** *If $a_m \to a_{m,0}$, as one might hope for if sample sizes converge to $\infty$,*

$$\Delta_v \to \mathbb{E}\left\{\sum_{k=1}^{K_v} \theta_0(k,v)\right\}.$$

*Note because $\theta_0(k,v) \geq 0$, all VIMP values are non-negative in the limiting case.*

Corollary 1 shows in the limit that each maximal $v$-subtree contributes equally to $\Delta_v$ through its node mean squared error. Furthermore, the node mean squared error for a subtree $\tilde{T}_{k,v}$ is a weighted average with some nodes contributing significantly more to overall mean squared error than others. From (4.3), we know that $\theta_0(k,v)$ is a weighted average involving weights $\pi_m$. The value $\pi_m$ is the probability that a fresh test $\mathbf{x}$ has node membership $m$ under $T$. If many splits are needed to reach $m$, or equivalently if $B_m(\mathbf{x})$ involves a large number of basis functions, then $\pi_m$ is likely to be small. Hence, nodes far down $\tilde{T}_{k,v}$ contribute less to $\theta_0(k,v)$ then nodes closer to its root node. Moreover, this effect is further amplified due to $\tilde{P}_{k,v}$. Under random splitting, if $L_m$ splits are needed to reach a node $m$ from the root of $\tilde{T}_{k,v}$, then $\tilde{P}_{k,v}$ assigns mass $2^{-L_m}$ to $a_m$. So again, the closer a node is to the root of the subtree, the bigger its impact on node mean squared error and $\Delta_v$. This is consistent with what was said earlier in Section 3 regarding impact node position has on VIMP.

## 5. Forests

In this section we extend our theory to forests. We begin with a formal definition.

DEFINITION 3. *Let $\mathscr{C} = \{T(\mathbf{x};b) : b = 1, \ldots, B\}$ be a collection of $1 \leq B < \infty$ binary trees grown from $\mathscr{L}$. The forest $\hat{\mu}_F$, for the collection of trees $\mathscr{C}$, is the ensemble predictor defined by*

$$\hat{\mu}_F(\mathbf{x}) = \frac{1}{B} \sum_{b=1}^{B} \hat{\mu}(\mathbf{x};b),$$

*where $\hat{\mu}(\mathbf{x};b)$ denotes the predictor for $T(\mathbf{x};b)$.*

Note unlike trees, forests do not assign node membership values. Forests are simply predictors, and they are much better at predicting than individual trees. In fact, under mild conditions, one can show that the closure of linear combination of binary tree predictors form a complete space. This leads to the following theorem (c.f. Proposition 2 of [13]).

**Theorem 2.** *Assume $\mathbf{x}$ has distribution $\mathbb{P}_0$ that is absolutely continuous with respect to Lebesgue measure with support $S_d$ on a finite closed rectangle in $\mathbb{R}^d$. Then for each real valued function $\mu \in L_2(\mathbb{P}_0)$, and each $\delta > 0$, there exists a forest $\hat{\mu}_F$ constructed from binary trees with $M > d$ terminal nodes such that*

$$\int_{S_d} (\hat{\mu}_F(\mathbf{x}) - \mu(\mathbf{x}))^2 \, \mathbb{P}_0(d\mathbf{x}) \leq \delta.$$

PROOF. The $L_2(\mathbb{P}_0)$-closure of finite linear combinations of indicator functions of rectangles over $S_d$ equals $L_2(\mathbb{P}_0)$. Therefore, if we can show that any indicator function for a rectangle on $S_d$ can be described by a predictor $\hat{\mu}(\mathbf{x})$ for a binary tree $T$ with $M > d$ nodes, the result follows because the closure of finite linear combinations of such predictors is $L_2(\mathbb{P}_0)$. It suffices to consider indicator functions of the form

$$I(\mathbf{x}) = I\{x_1 \le c_1, x_2 \le c_2, \ldots, x_d \le c_d\}.$$

To construct $T$, first split $x_1$ at $c_1$. This yields a left daughter node $x_1 \le c_1$. Now split the left daughter at $x_2 \le c_2$ and it's left daughter at $x_3 \le c_3$ and so forth up to $x_d \le c_d$. Label the resulting terminal node $m = 1$ and denote its basis function by $B_1(\mathbf{x})$. Observe that $d$ splits are needed to create $B_1(\mathbf{x})$ and therefore $T$ has $M = d + 1$ terminal nodes. Let $a_1 = 1$ and set $a_m = 0$ for $m > 1$. Then $\hat{\mu}(\mathbf{x}) = I(\mathbf{x})$. □

### 5.1. VIMP for forests

Suppose that $T(\mathbf{x}; b)$, $b = 1, \ldots, B$, are distinct binary trees grown from the learning data $\mathscr{L}$. For example, each tree could be grown as in random forests but without bootstrapping the data. Extending our previous notation in an obvious way, the forest predictor for the collection of trees is defined to be

$$\hat{\mu}_F(\mathbf{x}) = \frac{1}{B} \sum_{b=1}^{B} \sum_{m=1}^{M^{(b)}} a_m^{(b)} B_m(\mathbf{x}; b).$$

In light of Theorem 2, it seems quite reasonable to assume that $\hat{\mu}_F(\mathbf{x})$ is a good approximate to the true signal so long as $M^{(b)} > d$. Thus if each tree in the forest is deep enough that they have at least $d + 1$ terminal nodes, we can assume, analogous to (4.1), that

$$\mu(\mathbf{x}) = \frac{1}{B} \sum_{b=1}^{B} \sum_{m=1}^{M^{(b)}} a_{m,0}^{(b)} B_m(\mathbf{x}; b). \tag{5.1}$$

Noising up a variable $x_v$ in a forest is straightforward. One simply noises up $x_v$ for each tree as before. Then one computes the forest predictor, $\hat{\mu}_{F_v}(\mathbf{x})$, by averaging over the corresponding perturbed predictors $\hat{\mu}_v(\mathbf{x}; b)$. Extending notation in an obvious manner, by Lemma 1 we have

$$\hat{\mu}_{F_v}(\mathbf{x}) = \frac{1}{B} \sum_{b=1}^{B} \left\{ \sum_{m \notin M_v^{(b)}} a_m^{(b)} B_m(\mathbf{x}; b) + \sum_{k=1}^{K_v^{(b)}} \tilde{a}_{k,v}^{(b)} I\left\{T(\mathbf{x}; b) \in M_{k,v}^{(b)}\right\} \right\}.$$

Now using a similar argument as in Section 3.3, the VIMP for $x_v$ is:

$$\Delta_{F_v} = \mathbb{E}\left(R_{F_v}(\mathbf{x})^2\right) - 2\mathbb{E}\left\{R_{F_v}(\mathbf{x})\left[\mu(\mathbf{x}) - \hat{\mu}_F(\mathbf{x})\right]\right\},$$

where

$$R_{F_v}(\mathbf{x}) = \frac{1}{B} \sum_{b=1}^{B} \sum_{k=1}^{K_v^{(b)}} \sum_{m \in M_{k,v}^{(b)}} \left(\tilde{a}_{k,v}^{(b)} - a_m^{(b)}\right) B_m(\mathbf{x}; b).$$

From this we arrive at the following result.

**Theorem 3.** *Let $R_{F_v,0}(\mathbf{x})$ be the function $R_{F_v}(\mathbf{x})$ in which $a_m^{(b)}$ is replaced with the value $a_{m,0}^{(b)}$. Assume (3.2) and (5.1) holds. If $a_m^{(b)} \to a_{m,0}^{(b)}$ for each $m$ and $b$, then*

$$\Delta_{F_v} \to \mathbb{E}\left(R_{F_v,0}(\mathbf{x})^2\right) \leq \mathbb{E}\left\{\frac{1}{B}\sum_{b=1}^{B}\sum_{k=1}^{K_v^{(b)}} \theta_0(k,v,b)\right\}, \qquad (5.2)$$

*where*

$$\theta_0(k,v,b) = \sum_{m \in M_{k,v}^{(b)}} \pi_{m,b} \tilde{P}_{k,v,0}^{(b)} \left(\tilde{a}_{k,v,0}^{(b)} - a_{m,0}^{(b)}\right)^2$$

*is the node mean squared error for the $k$th maximal $v$-subtree of $T(\mathbf{x}; b)$ and $\pi_{m,b} = \mathbb{P}\{B_m(\mathbf{x};b) = 1 | \mathscr{L}\}$.*

PROOF. The limit follows automatically. The bound on the right-hand side of (5.2) is due to Jensen's inequality and orthogonality (2.1). □

### 5.2. Implications for random forests

The bound in (5.2) becomes tighter as the trees in the forest become orthogonal to one another. Moreover, a key assumption in Theorem (3) is that (5.1) holds, which by Theorem 2 is reasonable so as long as the individual trees in the forest are rich enough. Thus, a prescription for VIMP to be positive and fully characterized by subtree node mean squared error is that the forest should be derived from trees rich enough to approximate $\mu(\mathbf{x})$ while simultaneously being nearly orthogonal to one another. This is quite interesting, because this is the same prescription needed to achieve low generalization errors in random forests (see Theorem 11.2 of [2]). So the right-hand side of (5.2) is a very reasonable starting point for understanding VIMP in random forests. In fact, in Section 6.1, we will show that node mean squared error terms like those appearing in (5.2) play a crucial role in understanding paired associations between variables in forests.

## 6. Paired importance values and associations

For the moment let us return to a study of single trees. We consider the joint VIMP of a pair of variables $(x_v, x_w)$ in a binary tree $T$. Let $t = (v, w)$ and write $x_t$ to indicate $(x_v, x_w)$. The paired VIMP for $x_t$ is the difference between prediction error when $x_t$ is noised up versus the prediction error otherwise.

Noising up a pair of variables is defined as follows. For a case $\mathbf{x}$, drop $\mathbf{x}$ down $T$, and if $\mathbf{x}$ enters the root node of a maximal $v$-subtree $\tilde{T}_{k,v}$, or a maximal $w$-subtree $\tilde{T}_{k,w}$, then initiate a random left-right path until a terminal node is reached. This will assign $\mathbf{x}$ a random terminal value $\tilde{a}_{k,t}$ with distribution $\tilde{P}_{k,t}$.

Similar to our perturbation for a single variable, one can think of the above randomization process as defining a random tree. We refer to this tree as $T_t$. Let

$$\tilde{T}_{1,t}, \ldots, \tilde{T}_{K_t,t}$$

be the set of maximal $v$ and $w$-subtrees of $T$. Observe that in some cases a maximal $v$-subtree can be a $w$-subtree and a maximal $w$-subtree can be a $v$-subtree. Thus, $\{\tilde{T}_{k,t}\}$ are a collection of not necessarily disjoint trees. To avoid overlap, we let

$$\{\tilde{T}_{i_k,v}, k = 1, \ldots, K_v^*\} \text{ and } \{\tilde{T}_{i_l,w}, l = 1, \ldots, K_w^*\} \quad (6.1)$$

denote the set of *distinct* maximal $v$ and $w$-subtrees of $T$. The list (6.1) is constructed such that in the case where a maximal $v$-subtree and $w$-subtree intersect, the larger of the two subtrees is used. In particular, note that $K_v^* \leq K_v$ and $K_w^* \leq K_w$ where $K_v$ and $K_w$ are the number of distinct maximal $v$-subtrees and $w$-subtrees, respectively. When $K_v^* = K_v$ and $K_w^* = K_w$, then the set of $v$ and $w$-maximal subtrees are disjoint and $v$ and $w$ are orthogonal variables. Otherwise, $v$ and $w$ are associated. We will say more about this shortly.

Similar to Lemma 1, one can show that $\hat{\mu}_t$, the predictor for $T_t$, must be of the form:

$$\hat{\mu}_t(\mathbf{x}) = \sum_{m \notin M_t} a_m B_m(\mathbf{x}) + \sum_{k \in \mathscr{K}(v)} \tilde{a}_{k,v} I\{T(\mathbf{x}) \in M_{k,v}\}$$
$$+ \sum_{l \in \mathscr{K}(w)} \tilde{a}_{l,w} I\{T(\mathbf{x}) \in M_{l,w}\}.$$

Here $\mathscr{K}(v)$ and $\mathscr{K}(w)$ are the set of indices $\{i_k : k = 1, \ldots, K_v^*\}$ and $\{i_l : l = 1, \ldots, K_w^*\}$, and $M_t$ is the set of all terminal nodes for (6.1). Note importantly that $\tilde{a}_{k,v}$ and $\tilde{a}_{l,v}$ have distribution $\tilde{P}_{k,v}$ and $\tilde{P}_{l,w}$.

Define the paired VIMP for $x_t$ as

$$\Delta_t = \mathscr{P}(\hat{\mu}_t) - \mathscr{P}(\hat{\mu}).$$

We compare $\Delta_t$ to a composite additive effect of the individual variables $v$ and $w$. This will allow us to identify pairs of variables whose "joint effect" differs from the sum of their individual components. Thus allowing identification of interesting paired assocations (for another interesting way to identify interactions see [14]).

The marginal importance effect for $v$ and $w$ under an additive effect paradigm is defined to be

$$\Delta_{v+w} = \Delta_v + \Delta_w.$$

The difference between the two sets of importance values is what we call $A_t$, the *measure of association between $x_v$ and $x_w$*:

$$A_t = \Delta_t - \Delta_{v+w} = \Delta_t - (\Delta_v + \Delta_w). \quad (6.2)$$

Using similar arguments as in Theorem 1, we obtain the following characterization of $A_t$.

**Theorem 4.** *If $a_m \to a_{m,0}$, then under the conditions of Theorem 1:*

$$A_t \to -\mathbb{E}\left\{\sum_{k \notin \mathscr{K}(v)} \theta_0(k,v) + \sum_{l \notin \mathscr{K}(w)} \theta_0(l,w)\right\}.$$

Theorem 4 shows in the limit that $A_t$ can never be positive. The reason for this is that $\Delta_v$ and $\Delta_w$ overcount maximal subtrees since a maximal $v$-subtree can be a subset of a maximal $w$-subtree and vice-versa. Consequently, $\Delta_v + \Delta_w$ can never be smaller than $\Delta_t$. If $v$ and $w$ are orthogonal, then there is no overlap between maximal $v$ and $w$-subtrees and no overcounting occurs. Hence, the above sums are null and $A_t = 0$ in the limit. On the other hand, if $v$ and $w$ are not orthogonal, then $A_t$ has a negative limit. The more negative, the more overlap there is between $v$ and $w$-subtrees, and the stronger the indication of an association between the two variables.

### 6.1. Paired association values for forests

Association between variables is a little more delicate when it comes to forests. We show that association values can not only be negative, but they can also be positive in forests.

Extending notation in an obvious manner, the forest predictor under a noised up $x_t$ is defined as

$$\hat{\mu}_{F_t}(\mathbf{x}) = \frac{1}{B} \sum_{b=1}^{B} \left\{ \sum_{m \notin M_t^{(b)}} a_m^{(b)} B_m(\mathbf{x}; b) + \sum_{k \in \mathcal{K}(v,b)} \tilde{a}_{k,v}^{(b)} I\left\{ T(\mathbf{x}; b) \in M_{k,v}^{(b)} \right\} \right.$$
$$\left. + \sum_{l \in \mathcal{K}(w,b)} \tilde{a}_{l,w}^{(b)} I\left\{ T(\mathbf{x}; b) \in M_{l,v}^{(b)} \right\} \right\}.$$

Moreover, it follows straightforwardly from our previous arguments that the VIMP for $x_t$ is

$$\Delta_{F_t} = \mathbb{E}\left(R_{F_t}(\mathbf{x})^2\right) - 2\mathbb{E}\left\{R_{F_t}(\mathbf{x})\left[\mu(\mathbf{x}) - \hat{\mu}_F(\mathbf{x})\right]\right\},$$

where

$$R_{F_t}(\mathbf{x}) = \frac{1}{B} \sum_{b=1}^{B} \left\{ \sum_{k \in \mathcal{K}(v,b)} \sum_{m \in M_{k,v}^{(b)}} \left(\tilde{a}_{k,v}^{(b)} - a_m^{(b)}\right) B_m(\mathbf{x}; b) \right.$$
$$\left. + \sum_{l \in \mathcal{K}(w,b)} \sum_{m \in M_{l,w}^{(b)}} \left(\tilde{a}_{l,w}^{(b)} - a_m^{(b)}\right) B_m(\mathbf{x}; b) \right\}.$$

From this we obtain the following characterization.

**Theorem 5.** *Let $R_{F_t,0}(\mathbf{x})$ be the function $R_{F_t}(\mathbf{x})$ in which $a_m^{(b)}$ is replaced with the value $a_{m,0}^{(b)}$. Likewise, let $R_{F_v,0}(\mathbf{x})$ and $R_{F_w,0}(\mathbf{x})$ be defined as in Theorem 3. Assume (3.2) and (5.1) holds. If $a_m^{(b)} \to a_{m,0}^{(b)}$ for each $m$ and $b$, then*

$$\begin{aligned} A_t &= \Delta_{F_t} - (\Delta_{F_v} + \Delta_{F_w}) \\ &\to \mathbb{E}\left(R_{F_t,0}(\mathbf{x})^2\right) - \left(\mathbb{E}\left(R_{F_v,0}(\mathbf{x})^2\right) + \mathbb{E}\left(R_{F_w,0}(\mathbf{x})^2\right)\right). \end{aligned} \quad (6.3)$$

Theorem 5 implies a delicate relationship for $A_t$. This involves the way $v$ and $w$ interact within a given tree as well as how the variables interact across trees in the forest. Consider, first, the extreme case when all basis functions in $R_{F_t}$ are mutually orthogonal. This implies a forest where $v$ and $w$-maximal subtrees are disjoint *across* trees. Then,

$$A_t \to -\mathbb{E}\left\{\frac{1}{B}\sum_{b=1}^{B}\left(\sum_{k\notin \mathcal{K}(v,b)}\theta_0(k,v,b)+\sum_{l\notin \mathcal{K}(w,b)}\theta_0(l,w,b)\right)\right\},$$

where $\theta_0(k,v,b)$ and $\theta_0(l,w,b)$ are node mean squared errors as defined in Theorem 3. Note that the limit on the right-hand side is similar to that in Theorem 4. The only difference here is that we are averaging node mean squared error terms over trees. Just as in Theorem 4 the limit of $A_t$ cannot be positive. Also, the size of $A_t$ depends upon the behavior of $v$ and $w$ within a tree. For example, if $v$ and $w$ are orthogonal, such that for any given tree all maximal $v$ and $w$-subtrees are disjoint, then $A_t = 0$ in the limit. In contrast, the more overlap between $v$ and $w$-subtrees within a given tree, the more maximal subtrees are overcounted, and the more negative $A_t$ becomes. So just like single trees, large negative values indicate a strong association.

On the other hand, if trees are correlated, then (6.3) can be positive. One example is as follows. Suppose that $v$ and $w$ are highly correlated and both are influential. Then for any given tree both variables are strong competitors to grow the tree. If one variable is selected early on in the tree growing process, this might preclude the other variable from being selected. Things might be so bad that we could have 50% of the trees grown using $v$ and 50% grown using $w$. So noising up $v$ or $w$ individually will only affect 50% of the trees. However, a joint noising up of variables perturbs all trees. Consequently, prediction will be substantially worse and we would expect $A_t$ to be positive. The next example is a good illustration of this.

## 7. Air pollution

For our first illustration we use the well known "air pollution" data set [15]. The variables are daily readings of various air quality values measured from May 1, 1973 to September 30, 1973 in the New York metropolitan area. Solar radiation in Langleys (Solar), average wind speed in miles per hour (Wind), maximum daily temperature in Fahrenheit (Temp), month of the year (Month) and day of month (Day) were all recorded. The outcome is the mean ozone value (Ozone), which was transformed by taking its cube-root.

We analyzed the data using random forests. Computations were carried out using the "randomForest" R-package [16]. In regression settings, randomForest computes VIMP by subtracting an out-of-bag mean squared error rate under a random permutation of a variable $x_v$ from an out-of-bag mean squared error rate under the original $x_v$. As we have discussed already, these values resemble $\Delta_v$, but are not exactly the same.

Since randomForest does not compute paired importance values we adopted the following strategy for (6.2). To compute an analog of $\Delta_t$, we randomly selected 63% of the data to act as training data and set the remaining 37% aside as test data (these values represent the average in-bag and out-of-bag sample size). A forest with 1000 trees was grown over the training data using an *mtry* value of 3 (see [16] for details about options for randomForest). Then over the test data we randomly permuted $x_v$ and $x_w$ and created a noised up test data containing the permuted $x_v$ and $x_w$ variables. Using the forest predictor from the training data we computed the mean squared error over the modified test data and then we computed the mean squared error over the unmodified (original) test data. We then took the difference between these two values. This was repeated 1000 times independently. The average over the 1000 replicates is our proxy for $\Delta_t$. Finally, in order to be consistent, we computed proxies for $\Delta_v$ and $\Delta_w$ in exactly the same fashion. That is, we created a test data with a permuted $x_v$ and compared its test mean squared error to the test mean squared error for the original test data. The same was done for $x_w$.

Table 1 presents our results. All results are averaged over replicates. The first column, "Paired", records the paired VIMP for each possible pairwise comparison; the second column lists the additive effects for a given pair of variables (the sum of their individual importance values); the third column lists the association value (the difference between the paired VIMP and the additive effects); and the last column is the association value divided by the out-of-bag mean squared error for the training data multiplied by 100. This is a standardized value for association expressed in percentages.

TABLE 1
*Paired association values using random forests for air pollution data. Values reported are averaged over 1000 independent replicates with each forest based on 1000 trees.*

|  | Paired | Additive | Association | Assoc/MSE |
|---|---|---|---|---|
| Temp:Wind | 0.706 | 0.600 | 0.106 | 11.351 |
| Temp:Solar | 0.590 | 0.529 | 0.061 | 6.532 |
| Temp:Month | 0.464 | 0.455 | 0.008 | 0.904 |
| Temp:Day | 0.465 | 0.462 | 0.003 | 0.298 |
| Wind:Solar | 0.232 | 0.214 | 0.017 | 2.465 |
| Wind:Month | 0.142 | 0.141 | 0.001 | 0.137 |
| Wind:Day | 0.150 | 0.148 | 0.002 | 0.328 |
| Solar:Month | 0.074 | 0.073 | 0.000 | 0.041 |
| Solar:Day | 0.084 | 0.082 | 0.002 | 0.486 |
| Month:Day | 0.006 | 0.006 | 0.000 | 0.046 |

Table 1 shows that the largest association value is between the variables Temp and Wind. Also interesting, is the second largest association value occurring for Solar and Temp. Solar radiation and temperature are known to be related, and the fairly large association value in Table 1 confirms this relationship. The next largest association value is between Wind and Solar, but this value is substantially smaller (as can be seen by the standardized value in the 4th column of the table). After this, association values drop off dramatically.

These results are based on random forests, whereas Theorem 5 applies to the VIMP under a special type of noising up process for forests. Thus, it is interesting to see how well the theory extrapolates. To consider this, we generated a conditional plot (coplot) of ozone values against solar radiation, conditioning on wind and temperature. See Figure 2. As seen, ozone increase with solar radiation, and the slope of the curves increase as temperature increases for any fixed range of wind speeds. This clearly demonstrates an association between solar radiation and temperature. Also, looking at ozone and radiation with respect to wind, the slope increases as wind increases at high temperatures, but this pattern does not apply at lower temperature values. This is clear evidence of a Temp:Wind association. These results are consistent with Table 1.

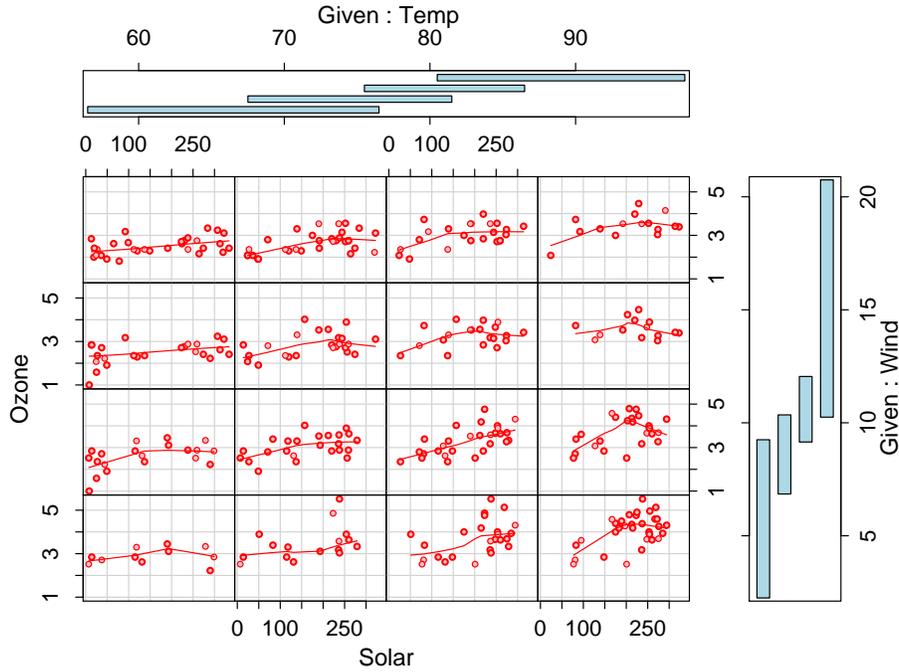

FIG 2. *Coplot of ozone values against solar radiation given wind and temperature.*

## 8. Simulation

A more direct confirmation of the theory can be illustrated using simulated data. For this example, data was simulated from the model

$$Y = 30\sin(\pi x_1 x_2) + 20(x_3 - 0.5)^2 + 20 x_1 x_4 + 5 x_5 + \varepsilon,$$

where $\varepsilon$ was distributed independently from a standard normal distribution. All variables $x_v$ were drawn from a uniform distribution on $[0,1]$. This included a 6th variable, $x_6$, representing noise, which was also part of the $x$-design matrix.

A sample size of $n = 100$ was drawn. The data was then analyzed using the randomForest package exactly as outlined in the previous section. All settings were kept the same. The simulation was repeated 100 times independently. The averaged values over the 100 simulations are reported in Table 2. Variables 1-6 are coded as a-f respectively in Table 2.

TABLE 2
*Paired association values using random forests for simulated data. Analysis carried out in the same manner as Table 1.*

|     | Paired  | Additive | Association | Assoc/MSE |
| --- | ------- | -------- | ----------- | --------- |
| a:b | 185.216 | 192.870  | −7.654      | −4.231    |
| a:c | 132.347 | 132.426  | −0.079      | −0.044    |
| a:d | 140.472 | 141.906  | −1.434      | −0.849    |
| a:e | 133.518 | 133.530  | −0.011      | −0.003    |
| a:f | 131.552 | 131.639  | −0.088      | −0.055    |
| b:c | 61.692  | 61.736   | −0.044      | −0.033    |
| b:d | 71.350  | 71.200   | 0.150       | 0.110     |
| b:e | 62.822  | 62.857   | −0.035      | −0.023    |
| b:f | 60.990  | 60.965   | 0.025       | 0.021     |
| c:d | 10.888  | 10.870   | 0.018       | 0.014     |
| c:e | 2.500   | 2.474    | 0.026       | 0.026     |
| c:f | 0.599   | 0.589    | 0.009       | 0.007     |
| d:e | 11.948  | 11.934   | 0.015       | 0.019     |
| d:f | 10.022  | 10.040   | −0.019      | −0.022    |
| e:f | 1.713   | 1.715    | −0.002      | −0.011    |

By far, the two largest standarized association values are a:b and a:d, represening the interactions between variable $x_1$ and $x_2$ and $x_1$ and $x_4$ (see the 4th column in Table 2). Note how these values are negative, unlike in our previous analysis where all significant associations were positive. Large negative values like these, as our theory suggests, is due to high overlap between $v$ and $w$-subtrees. For example, for $x_1$ and $x_2$, their interaction must be such that whenever $x_1$ splits a node in the tree, there is a high likelihood that $x_2$ is split underneath this node, and vice-versa.